\documentclass[10pt,twocolumn,letterpaper]{article}

\usepackage{cvpr}
\usepackage{times}
\usepackage{epsfig}
\usepackage{graphicx}
\usepackage{amsmath}
\usepackage{amssymb}
\usepackage{subfigure}
\usepackage{setspace}

\usepackage[breaklinks=true,bookmarks=false]{hyperref}

\cvprfinalcopy 

\def\cvprPaperID{****} 

\begin{document}

\title{Fully Convolutional Model for Variable Bit Length and Lossy High Density Compression of Mammograms}

\author{Aupendu Kar, Sri Phani Krishna Karri, Nirmalya Ghosh, Debdoot Sheet\thanks{This work is supported under the Intel India Grand Challenge 2016 grant for Project MIRIAD.}\\
Department of Electrical Engineering, Indian Institute of Technology Kharagpur\\
Kharagpur, West Bengal, India\\
{\tt\small debdoot@ee.iitkgp.ac.in}
\and
Ramanathan Sethuraman\\
Intel Technology India Pvt. Ltd.\\
Bangalore, Karnataka, India\\
{\tt\small ramanathan.sethuraman@intel.com}
}

\maketitle

\begin{abstract}

Early works on medical image compression date to the 1980's with the impetus on deployment of teleradiology systems for high-resolution digital X-ray detectors. Commercially deployed systems during the period could compress 4,096 $\times$ 4,096 sized images at 12 bpp to 2 bpp using lossless arithmetic coding, and over the years JPEG and JPEG2000 were imbibed reaching upto 0.1 bpp. Inspired by the reprise of deep learning based compression for natural images over the last two years, we propose a fully convolutional autoencoder for diagnostically relevant feature preserving lossy compression. This is followed by leveraging arithmetic coding for encapsulating high redundancy of features for further high-density code packing leading to variable bit length. We demonstrate performance on two different publicly available digital mammography datasets using peak signal-to-noise ratio (pSNR), structural similarity (SSIM) index and domain adaptability tests between datasets. At high density compression factors of $>$300$\times$ (~0.04 bpp), our approach rivals JPEG and JPEG2000 as evaluated through a Radiologist's visual Turing test.

\end{abstract}

\section{Introduction}

\begin{figure}[t]
    \centering
       \includegraphics[width=0.45\textwidth]{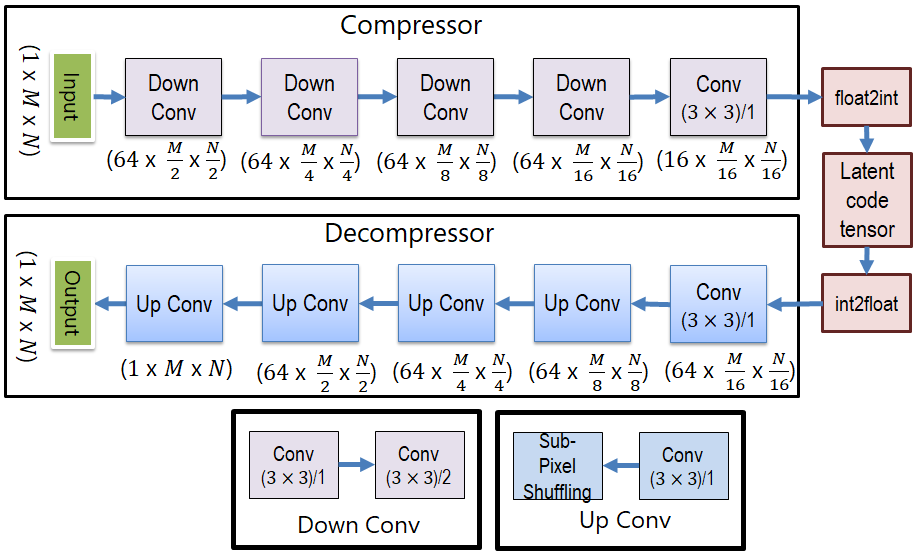}
  \caption{Overview of the convolutional autoencoder (CAE) with adaptive arithmetic encoding of the latent code tensor for variable bit length mammogram compression.}
  \label{graphicalabstract} 
\end{figure}

While image and video compression for consumer grade cameras have been in use since early 1970's, it was more than two decades later, with the advent of television and telconferencing, that medical images started being compressed. Availability of digital X-ray detector, development of full-scale digital imaging systems, deployment of teleradiology networks for screening, life long archival of medical images for pathology modeling and personalized medicine were few of the factors that inspired early research on medical image compression in 1980's. These were deployed a decade later~\cite{kuduvalli1992performance} for addressing the challenges faced by clinical establishments. Hospitals typically accumulate about 4TB of imaging data per year, and with the growing trend towards personalized medicine necessitating life long archival, it inspires development of high-density medical image compression at factors $>300\times$ achieving $<0.05$ bpp with no loss of visual features relevant for diagnosis. Our method in Fig.~\ref{graphicalabstract} is inspired by recent developments in deep learning based compression techniques for camera images~\cite{toderici2016full, johnston2017improved, theis2017lossy, rippel2017real} and experimentally rivals JPEG ~\cite{wallace1992jpeg} as well as the reigning standard JPEG2000 ~\cite{skodras2001jpeg} at such demands of high compression factors, validated with experiments performed using X-ray mammograms.

The rest of the paper is organized as follows. Prior art is discussed in Sec. 2. Sec. 3 describes the fully convolutional autoencoder (CAE) based compression engine for mammograms. The experiments performed for evaluation and benchmarking of performance of our approach vs. JPEG and JPEG2000 are described in Sec. 4 and characteristic aspects along with clinical usability study is discussed in Sec. 5, finally concluding the work in Sec. 6.

\section{Related work}

Information compression in digital media including images can typically be grouped into lossless or lossy. Entropy based methods, also known as arithmetic encoding techniques are popular for lossless compression~\cite{kuduvalli1992performance} and on the other hand transformed domain compression using discrete consine transform (DCT) based JPEG~\cite{wallace1992jpeg} and discrete wavelet transform (DWT) based JPEG2000~\cite{skodras2001jpeg} are the reigning popular standards for lossy image compression~\cite{saha2000image}. Radiological image compression since 1980's has been lossless at its infancy~\cite{wong1995radiologic} to avoid loss of any potential features of diagnostic relevance~\cite{kuduvalli1992performance}. With gain in pixel-density of sensors and availability of high-resolution full-scale radiological scanners, JPEG and JPEG2000 have seen entry for radiological image compression, especially mammograms~\cite{khademi2006comparison}. In view of the distortions typical to JPEG based compression~\cite{corchs2014no}, limiting its use for high-density image compression, recent developments in learning based approaches have been proposed to ensure ability to learn to preserve representative structures in images. Recent efforts for color image compression use recurrent convolutional neural network \cite{toderici2016full,johnston2017improved}. Subsequently fully convolutional architectures without any recurrence \cite{theis2017lossy} have also outperformed models with recurrence. Recent developments also include adversarial learning \cite{rippel2017real} to achieve visually smooth decompression of color images. We address the aspect of learning based medical image compression, especially in mammograms, to develop fully convolutional neural network based methods to overcome limitations of related prior-art~\cite{tan2011using}.

\section{Methodology}

Our model for fully convolutional image compression consists of (a) compressor-decompressor blocks trained as a convolutional autoencoder, (b) adaptive arithmetic encoding for further lossless compression of the bit-length.

\textbf{Compressor-Decompressor:} The compressor is learned as the encoder unit and the decompressor as the decoder unit of a fully convolutional autoencoder. Fig.~\ref{graphicalabstract} shows the architecture of the autoencoder used here. The \emph{compressor} consists of down-convolution (down-conv) blocks to extract key features and reduce the bits allocated for storage. The symmetrically shape matched up-convolution (up-conv) blocks in the \emph{decompressor} reconstruct the image from the compressed bitstream. 

\emph{Compressor:} Each down-conv block consists of a convolution layer with $3\times3$ kernel and stride of 1 followed by ReLU() activation, and subsequently a second convolution layer with $3\times3$ kernel and stride of 2 followed by ReLU() activation. Mammograms typically have bit-depths ranging 12-/16 bpp and for our purpose they are range normalized in $[0,1]$ represented in floating point tensors. The input image is processed through 4 stages of down-conv, followed by a convolution layer with $3\times3$ kernel and stride of 1, with ClippedReLU() activation function with clipping at 1. 

\emph{Latent code tensor:} This is generated using the $float2int()$ operation defined as $g(x,y,c)=\left[(2^n-1)i(x,y,c)\right]$ where $i(x,y,c)\in[0,1]$ is the floating point value obtained from the compressor, $[\cdot]$ is the integer rounding off operator, $n$ is the bit-length of each element $g(x,y,c)$ in the latent code tensor, and $x,y,c$ correspond to the spatial and channel index specifying location of the scalar in the compressed code tensor. The $int2float()$ operator converts the integer valued latent code tensor to floating point value $j(x,y,c)=(2^n-1)^{-1}g(x,y,c)$. These being non-differentiable, $\nabla float2int()$ and $\nabla int2float()$ are approximated as 1.

\emph{Decompressor:} The first convolution layer consists of $3\times 3$ sized kernels with stride of 1 and ReLU() activation function, followed by 4 units of up-conv blocks. Each up-conv block consists of $3\times 3$ sized convolution kernels with stride of 1 and ReLU() activation followed by sub-pixel shuffling~\cite{shi2016real}. The last up-conv block uses a ClippedReLU() with clipping at 1 as activation function instead of ReLU().

\textbf{Adaptive arithmetic encoding:} This phase comes into action only during deployment and not during training. Here the integer value latent code tensor representing the compressed version of the image is linearized into a 1-D array following either row-major or column-major representation, such that each element is a n-bit long integer. Hence a tensor of size $k\times m \times c$ would be a $kmcn$ bit-long representation. Subsequent to this, 8-bit long bit streams are extracted to have $kmcn/8$ codes that are compressed losslessly following entropy based adaptive arithmetic encoding~\cite{witten1987arithmetic}. This stage further compresses the bit-stream, on account of the high amount of spatial redundancy in mammograms to obtain high-density compression.

\section{Experiments and Results}

\textbf{Dataset description}: \emph{CBIS-DDSM}\footnote{https://wiki.cancerimagingarchive.net/display/Public/CBIS-DDSM/} and \emph{Dream}\footnote{https://www.synapse.org/\#!Synapse:syn4224222} are the two publicly available databases of digital mammogram images that are used for training and performance validation of the compression engines. Mammograms in \emph{Dream} are encoded at $12$ bpp and the ones in \emph{CBIS-DDSM} are encoded either as $16$ bpp or $8$ bpp. In experiments with \emph{CBIS-DDSM} only $16$ bpp are used since $8$ bpp is typically not employed for digital mammography during acquisition. Out of the $3,102$ mammograms in $16$ bpp, a subset of $102$ randomly selected ones are used for testing. In experiments with \emph{Dream} $480$ mammograms are used for training the model and $20$ for testing.

\textbf{Training:} While trainig the \emph{compressor-decompressor} jointly, $256\times256$ sized randomly located patches from the range normalized mammogram are used. A patch is included in training set only when $>50\%$ pixels are non-zero valued, the mean intensity of the patch is not $0$ or $1$, and variance $>0$. In \emph{Dream} we use $3,840$ patches and in \emph{CBIS-DDSM} we use $3,000$ patches.

\textbf{Training parameters:} The Adam optimizer \cite{kingma2014adam} with $\epsilon = 10^{-8}$, $\beta_{1} = 0.9$, $\beta_{2} = 0.999$ is used. All models are trained over $1,000$ epochs with a learning rate of $10^{-4}$ and the batch size of $16$. Mean squared error (MSE) is used as the loss function for back propagating error during training.

\textbf{Baselines:} We compare the performance of our approach against conventional methods like JPEG, JPEG2000. We also compare against a fully-connected autoencoder (FCAE) model inspired from~\cite{tan2011using} and optimized for high-density compression. 

FCAE is trained using $16\times16$ sized patches collected following the inclusion principle used for CAE training. The \emph{compressor} with $N=16^2$ connects $N\rightarrow 8N \rightarrow 4N \rightarrow 2N$ hidden neuron with tanh() activation, and to $N$ binary neurons representing the \emph{latent code tensor}. The \emph{decompressor} consists of $N \rightarrow 2N \rightarrow 4N \rightarrow 8N$ hidden neuron with tanh() activation, and to $N$ neurons with ClippedReLU() activation function with clipping at 1 to obtain the decompressed patch sized $16 \times 16$. During compression mammograms are divided into non-overlapping sequential blocks, and decompressor reorders it for retrieval.

\textbf{Implementation:} Both CAE and FCAE models are implemented with PyTorch on Python 2.7, accelerated with CUDA 9.0 on Nvidia Quadro P6000 with 24 GB DDR5 RAM on a PC with Intel Core i5 CPU and 28 GB of system RAM running Ubuntu 16.04 LTS. Average training time is $\sim 30$ sec / epoch. The compressor has $268,368$ learnable parameters and decompressor has $454,724$. 

\begin{figure}[t]
    \centering
    \subfigure[Original, 12 bpp]
    {
        \includegraphics[width=0.2\textwidth, trim={800 1938 2100 1760}, clip]{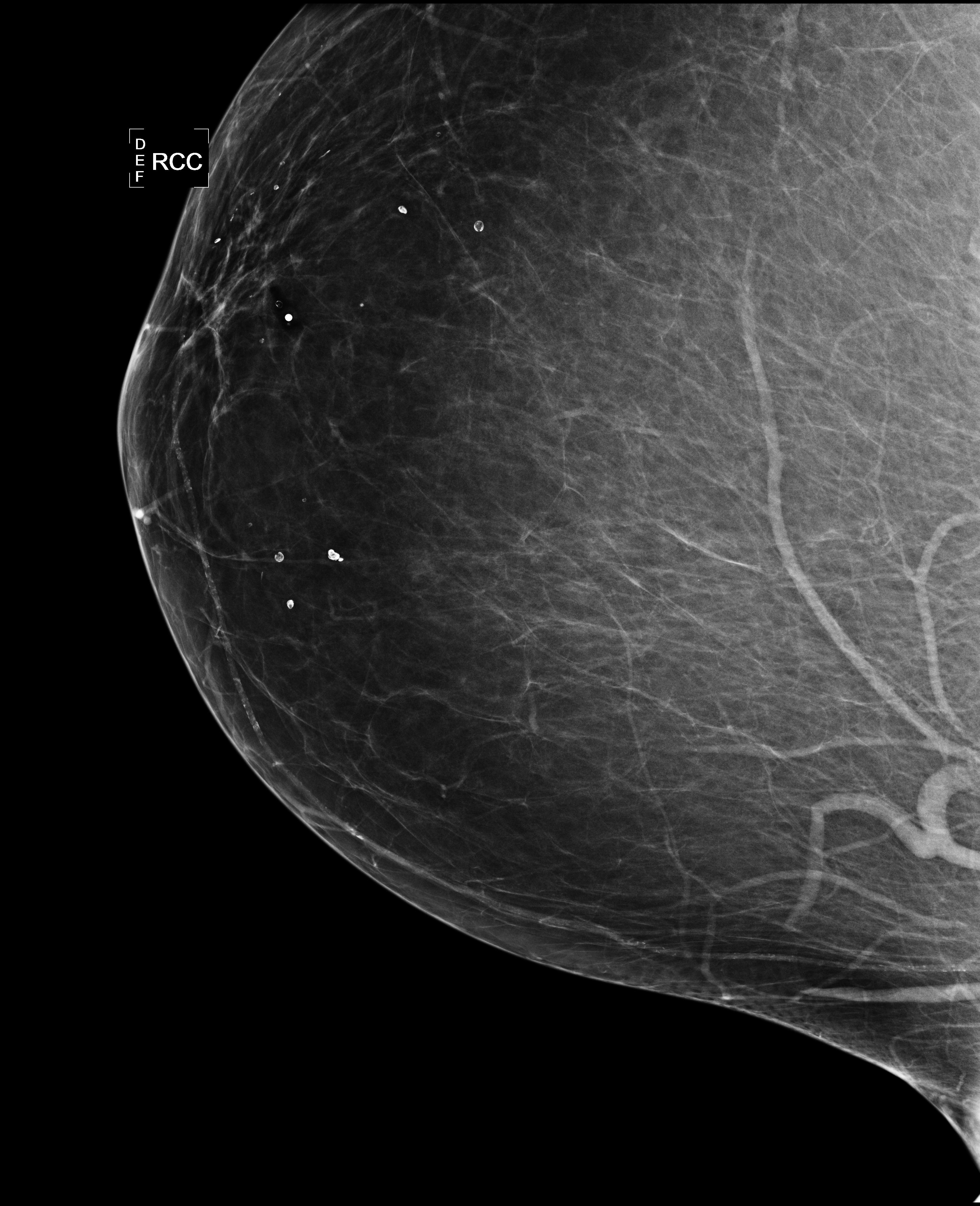}
        \label{vieworig}
    }
    \subfigure[JPEG, 0.132 bpp]
    {
        \includegraphics[width=0.2\textwidth, trim={800 1938 2100 1760}, clip]{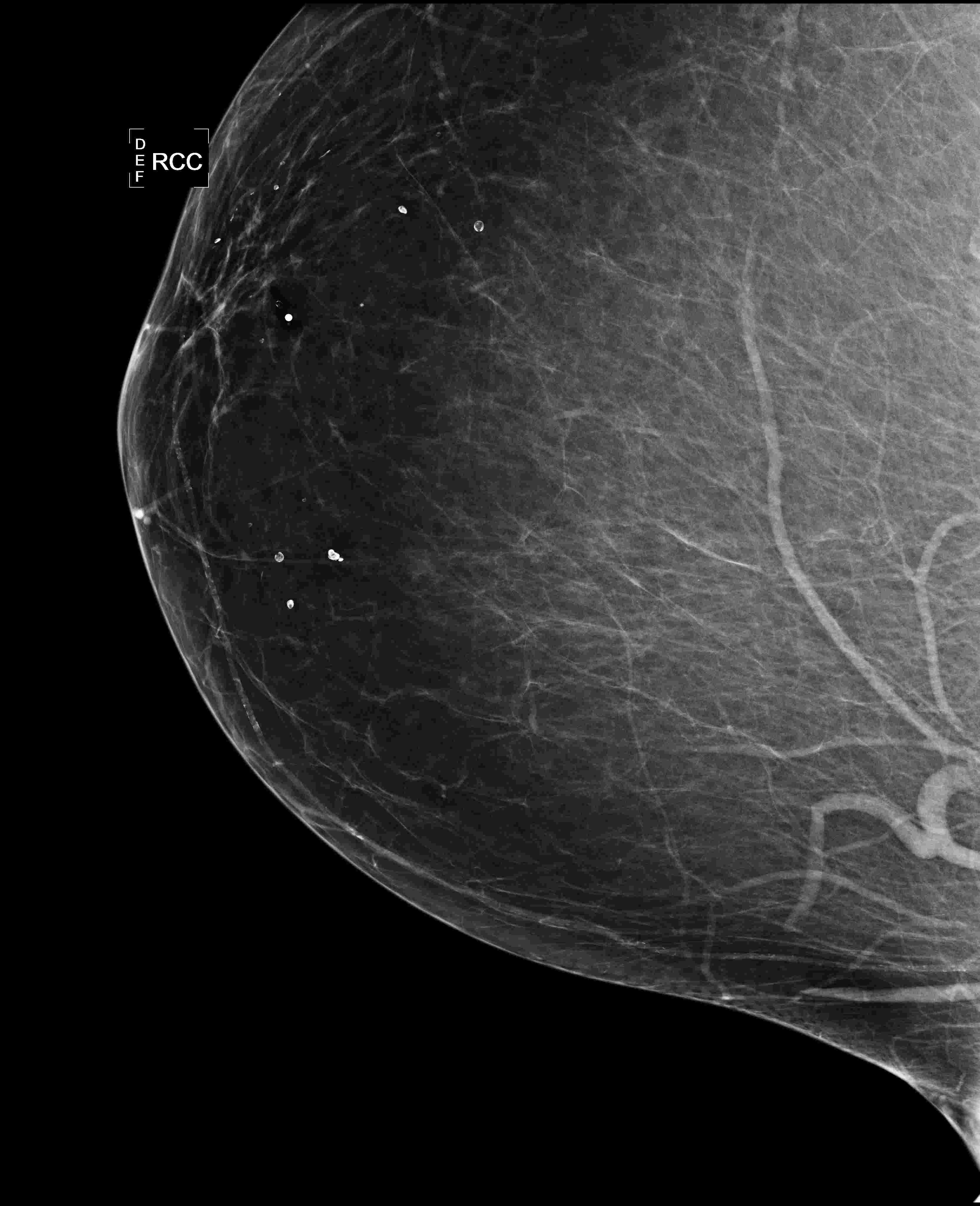}
        \label{viewjpg}
    }
    \subfigure[JPEG2000, 0.051 bpp]
    {
        \includegraphics[width=0.2\textwidth, trim={800 1938 2100 1760}, clip]{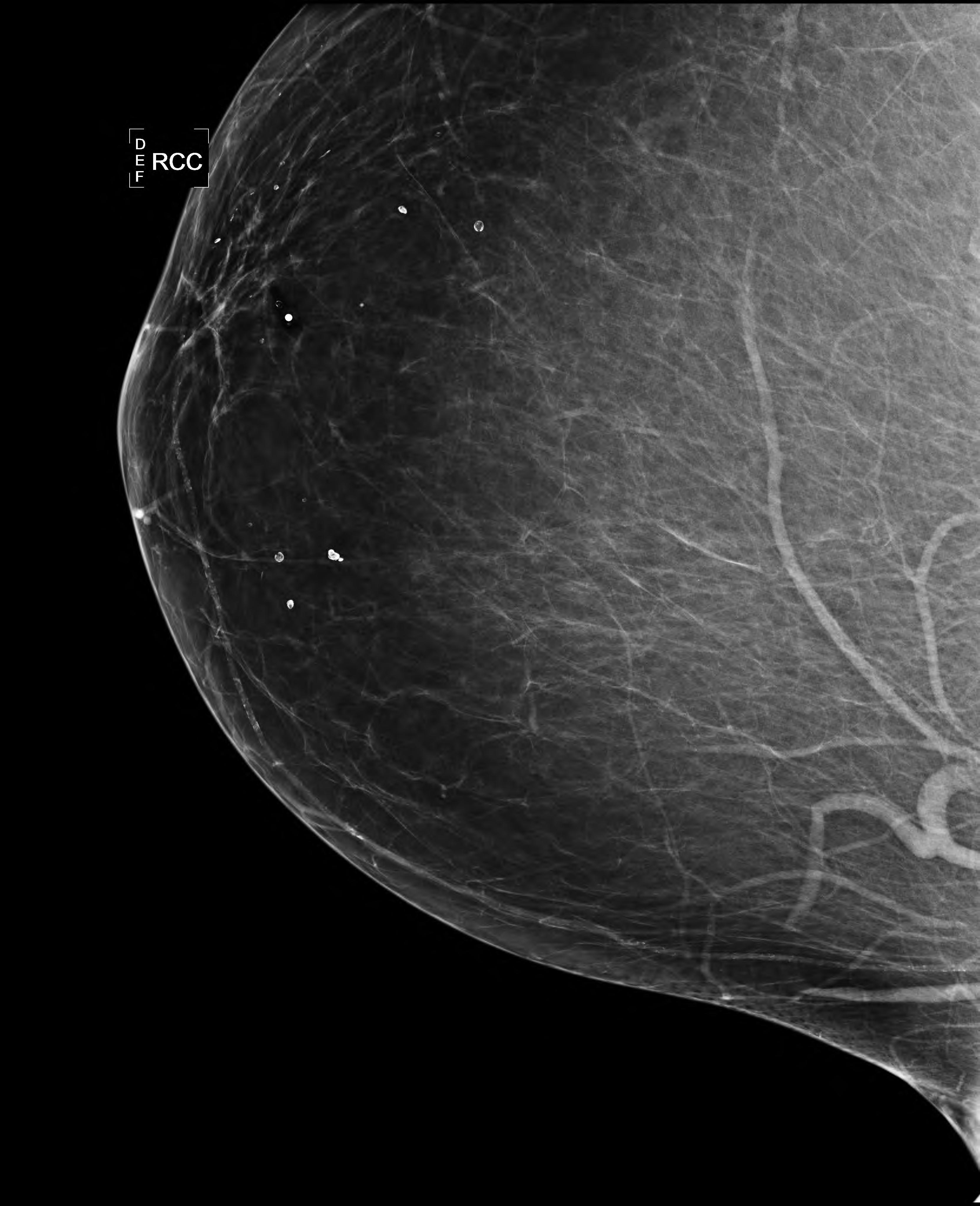}
        \label{viewjp2}
    }
    \subfigure[CAE, 0.049 bpp]
    {
        \includegraphics[width=0.2\textwidth, trim={800 1938 2100 1760}, clip]{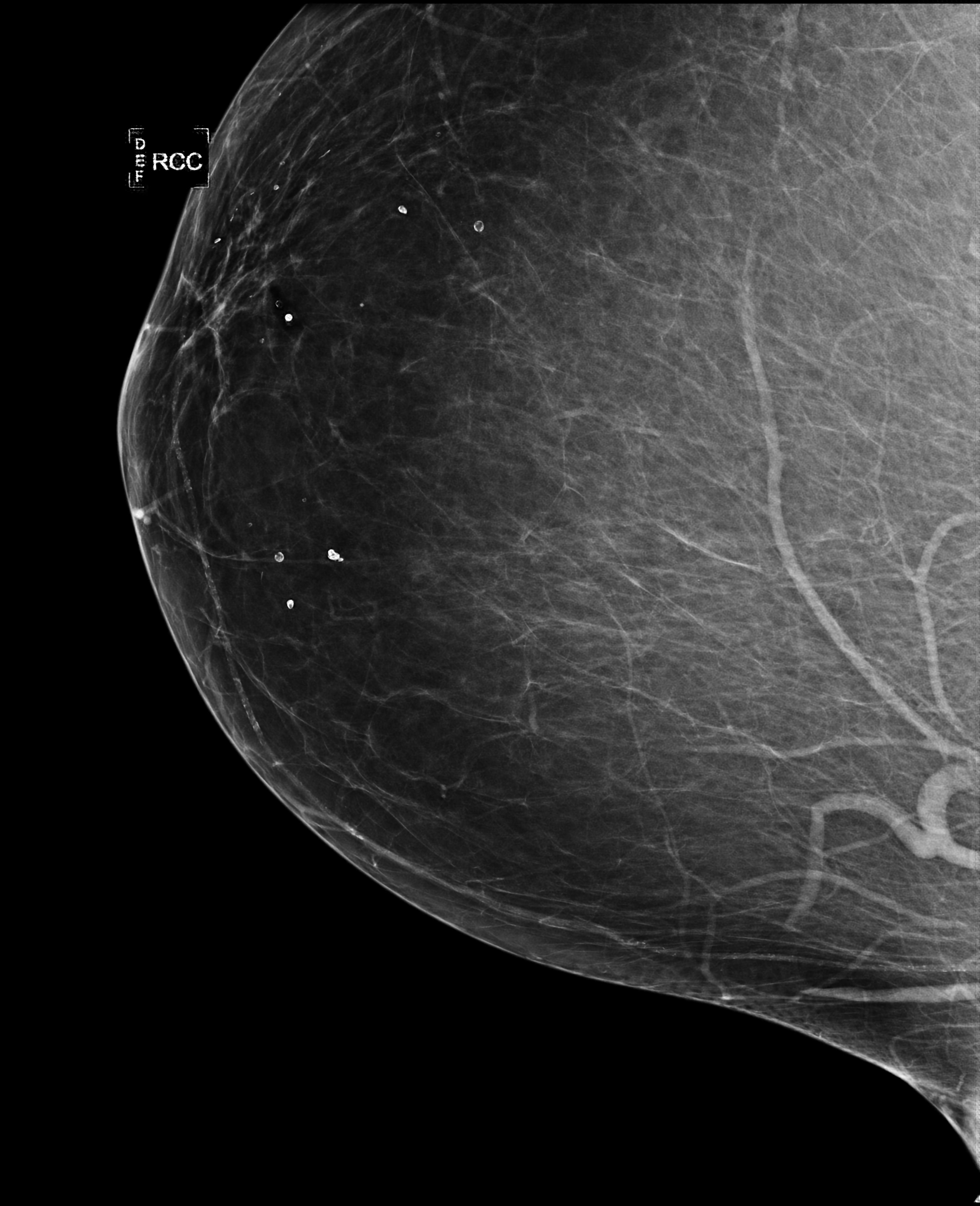}
        \label{viewcae}
    }
    \caption{Calcified disc on a sample from Dream.}
    \label{view}
\end{figure}

\begin{figure}[t]
    \centering
    \subfigure[Tested on CBIS-DDSM]
    {
        \includegraphics[width=0.45\textwidth]{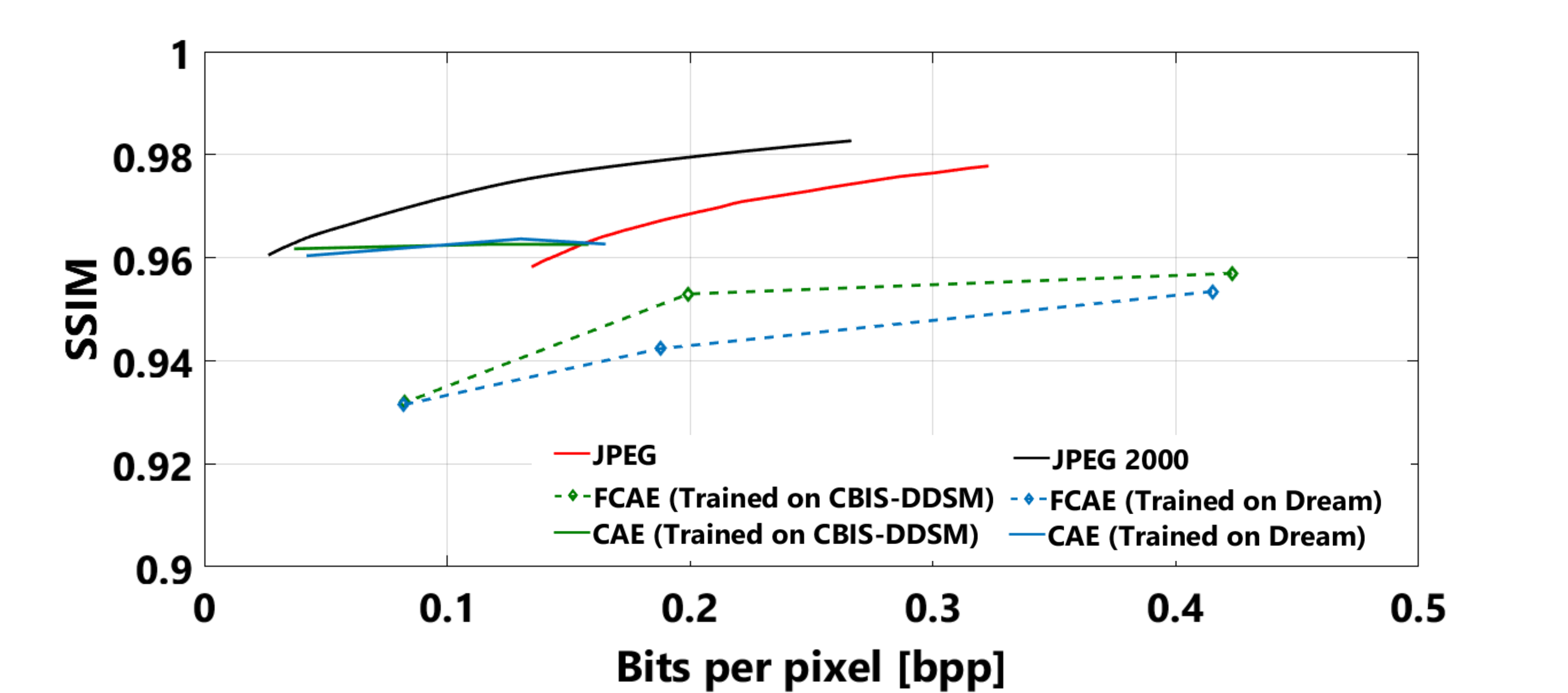}
        \label{caeplota}
    }
    \\
    \subfigure[Tested on CBIS-DDSM]
    {
        \includegraphics[width=0.45\textwidth]{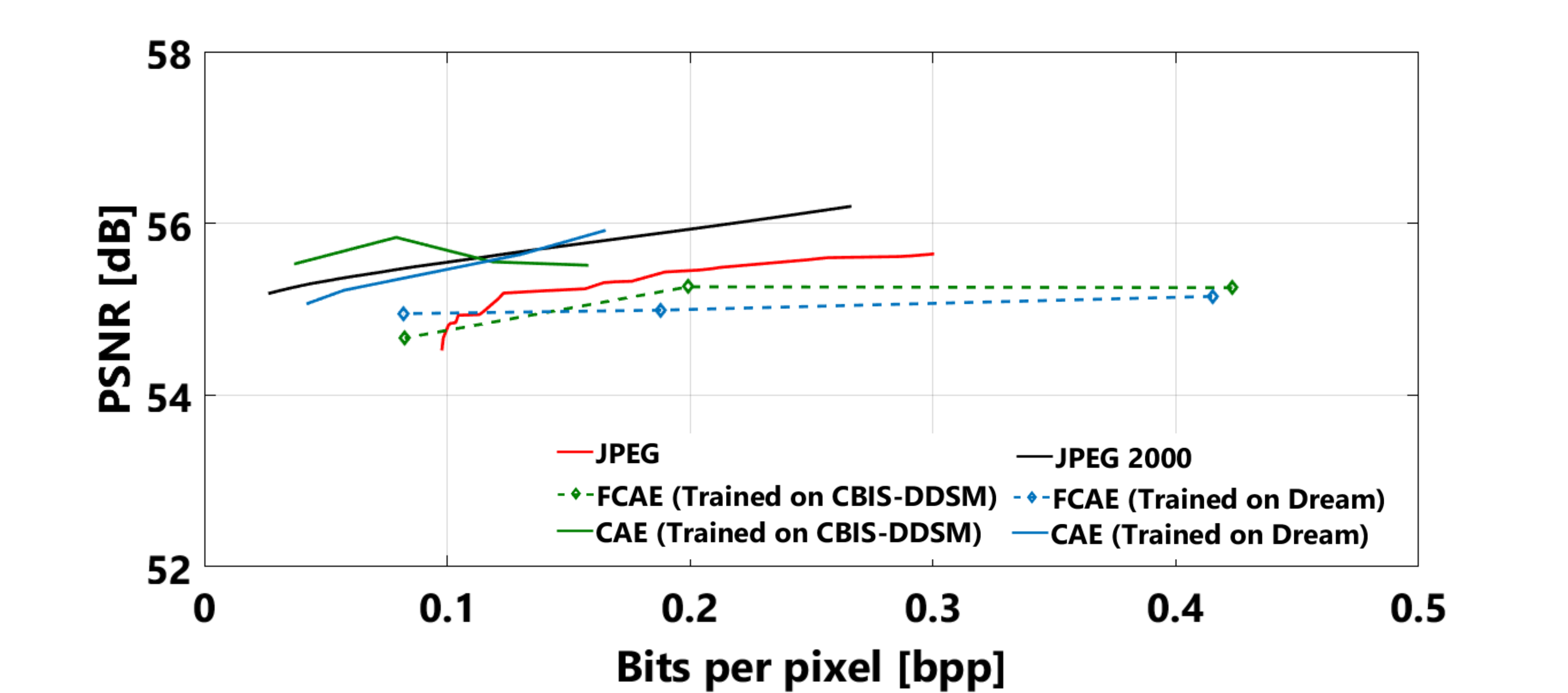}
        \label{caeplotb}
    }
    \subfigure[Tested on Dream]
    {
        \includegraphics[width=0.45\textwidth]{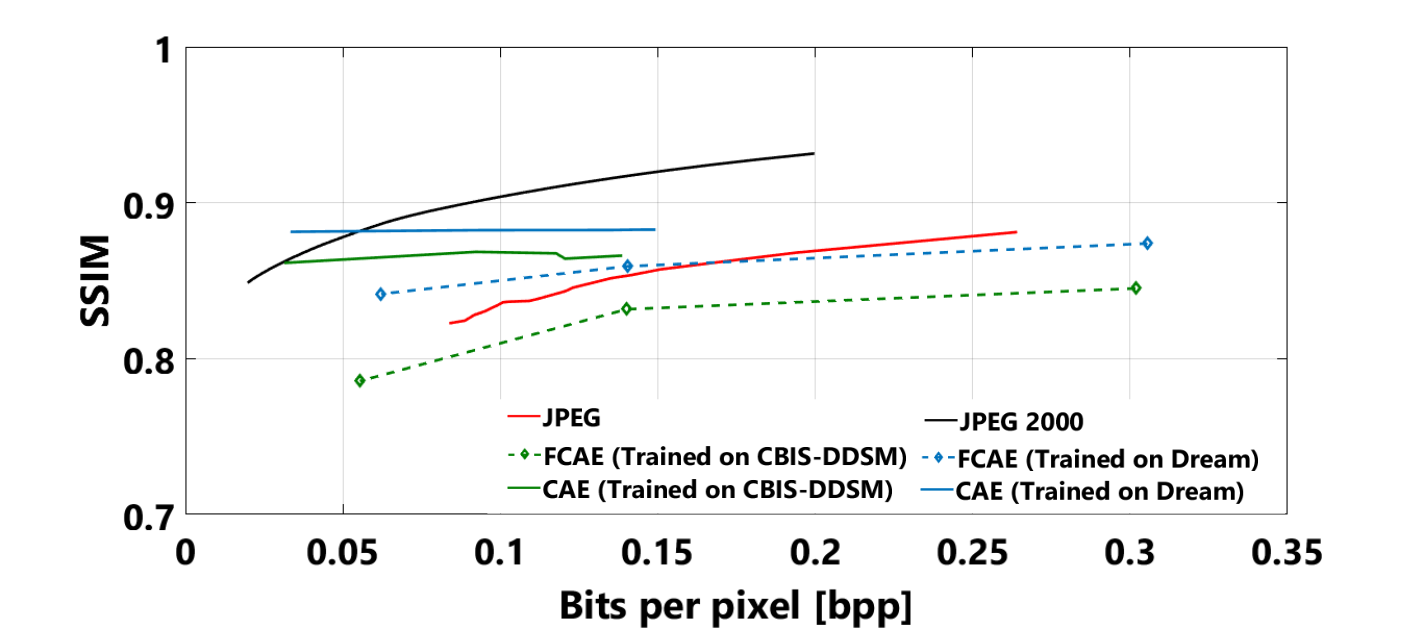}
        \label{caeplotc}
    }
    \subfigure[Tested on Dream]
    {
        \includegraphics[width=0.45\textwidth]{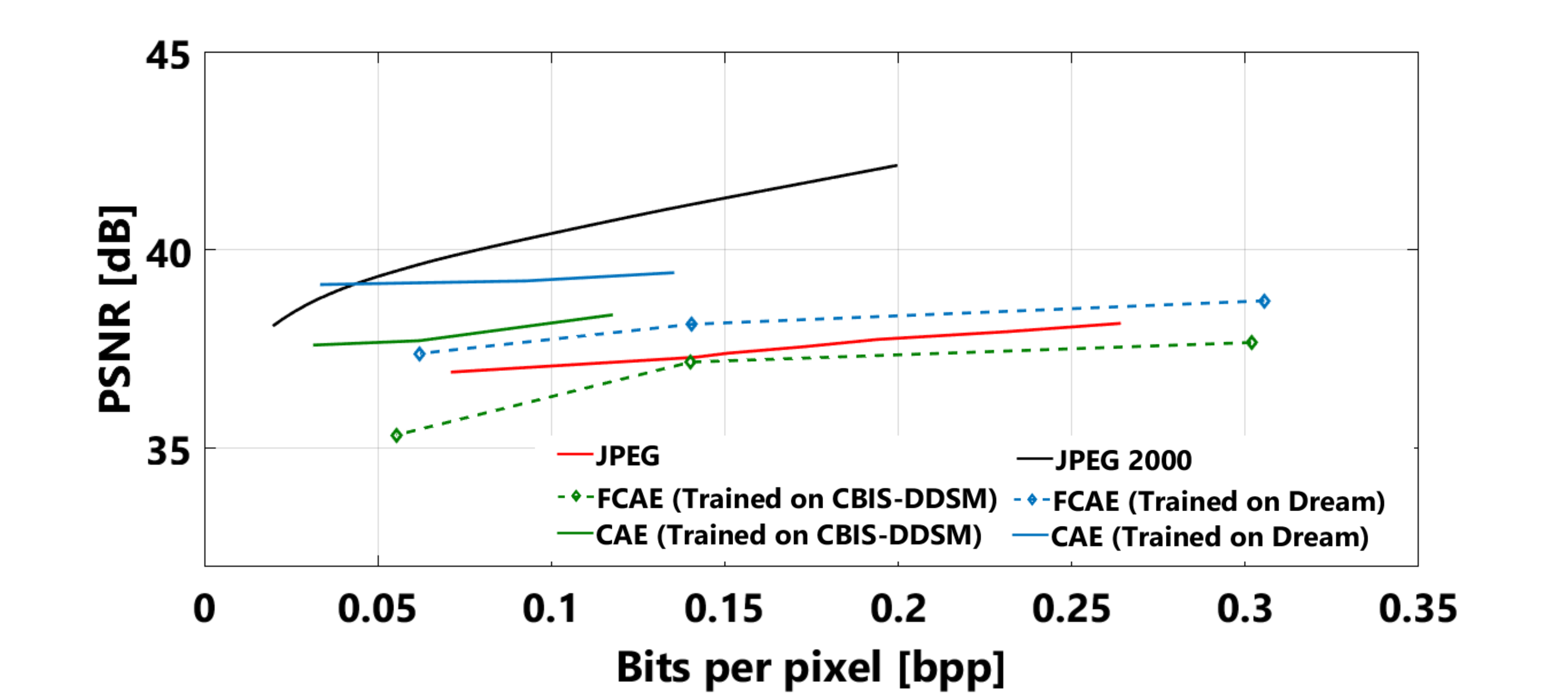}
        \label{caeplotd}
    }
    \caption{Performance comparison of JPEG, JPEG2000 and FCAE with CAE for mammogram compression, trained and evaluated on \emph{Dream} and \emph{CBIS-DDSM}. The bpp and compression factors are factored from the effective file size following adaptive arithmetic encoding of the latent code tensor. pSNR and SSIM for FCAE could be calculated only at some fixed compression factors and not across the whole range owing to the dependency of compression factor on the architecture.}
    \label{caeplot}
\end{figure}

\textbf{Results:} Fig.~\ref{view} presents qualitative results while Fig.~\ref{caeplot} shows the pSNR and SSIM \cite{wang2004image} plots at different bpp corresponding to different compression factors on the two datasets used for different compression engines. While Figs.~\ref{caeplota}, \ref{caeplotb} show the results of testing it on \emph{CBIS-DDSM}, Figs.~\ref{caeplotc}, \ref{caeplotd} show tests on \emph{Dream}. While the focus of high-density image compression is to be domain specific, we specifically evaluate performance of the method on \emph{Dream} when trained with \emph{CBIS-DDSM} and vice-versa as well, to observe interesting trends on their cross-dataset generalizability in the same domain.

\section{Discussions}

\textbf{Quantitative evaluation} of the CAE based compression engine outperforms JPEG and JPEG2000 at higher compression factors yielding $<0.1$ bpp in terms of both pSNR and SSIM. The fact that CAE has a relatively flat value of image quality over a wide dynamic range of bpp is noteworthy. The variation in bpp is brought in by varying $n$ in the \emph{latent code tensor}. It was observed that the values of $g(x,y,c)$ in the \emph{latent code tensor} typically do not densely range over the complete range of $[0,2^n-1]$ but only over a smaller set of values occupying some $k<<2^n$ number of possible values; thus exhibiting a relatively similar value of entropy $(H)$ until a value of $n$ lower than $H$, e.g. $H=2.95,2.51, 2.44, 1.98$ at $n=14, 10, 6, 2$ respectively. This corroborates with a relatively flat response in pSNR and SSIM for CAE, while those for FCAE, JPEG and JPEG2000 steadily decrease with lowering of bpp. Observing in Fig.~\ref{view} it is evident that a given SSIM mark of $0.80$ can be achieved with FCAE at $0.101$ bpp, CAE at $0.049$ bpp, JPEG2000 at $0.051$ bpp and JPEG at $0.132$ bpp.

\textbf{Visual Turing test} (VTT) was performed where $5$ Radiologist's were asked to visually inspect and identify the uncompressed image out of a pair where the other image was its corresponding compressed version. 10 mammograms were used in each test. With CAE $50\%$ of images could be identified as uncompressed vs. $52\%$ with JPEG2000.

\section{Conclusion}

We have proposed a fully convolutional approach for high-density compression of mammograms without loss of diagnostically relevant  pathological features. The use of entropy based lossless arithmetic encoding further boosts compression factor, and our CAE's ability to represent images in a sparse code space leads to a relatively flat image quality over a wide range of compression factors. Contrary to JPEG producing artifacts on decompression, our approach does not introduce such distortions. Visual scoring of results by reviewing Radiologists have reported superiority on diagnostic relevance over JPEG2000 as well.

{\small
\bibliographystyle{ieee}
\bibliography{egbib}
}

\end{document}